\documentclass{article}
\usepackage{spconf,amsmath,graphicx}
\usepackage{multirow}
\usepackage{hyperref}
\usepackage{subcaption}
\usepackage{xcolor}
\usepackage{slashbox}
\usepackage{adjustbox}
\usepackage{makecell}
\usepackage{diagbox}
\usepackage{setspace}
\usepackage{amssymb}
\usepackage{mathtools}
\usepackage[T1]{fontenc}
\newcommand\tab[1][1cm]{\hspace*{#1}}
\DeclareCaptionLabelFormat{andtable}{\textbf{\tablename~\thetable~\&  #1~#2}}
\hypersetup{ colorlinks=false, linkcolor=blue, filecolor=black, urlcolor=blue, }

\title{RANDOMSEMO: NORMALITY LEARNING OF MOVING OBJECTS \\FOR VIDEO ANOMALY DETECTION}
\name{Chaewon Park\sthanks{E-mail: chaewon28@yonsei.ac.kr}, Minhyeok Lee\sthanks{E-mail: hydragon516@yonsei.ac.kr}, MyeongAh Cho\sthanks{E-mail: maycho0305@yonsei.ac.kr} Sangyoun Lee\sthanks{Corresponding Author, E-mail: syleee@yonsei.ac.kr}}
\address{School of Electrical and Electronic Engineering, Yonsei University, Seoul, Korea}

\usepackage{verbatim}
\begin{document}
%\ninept
%
\maketitle
\begin{abstract}
	Recent anomaly detection algorithms have shown powerful performance by adopting frame predicting autoencoders. 
	However, these methods face two challenging circumstances. 
	First, they are likely to be trained to be excessively powerful, generating even abnormal frames well, which leads to failure in detecting anomalies.
	Second, they are distracted by the large number of objects captured in both foreground and background. 
	To solve these problems, we propose a novel superpixel-based video data transformation technique named Random Superpixel Erasing on Moving Objects (\textit{RandomSEMO}) and Moving Object Loss (\textit{MOLoss}), built on top of a simple lightweight autoencoder. 
	\textit{RandomSEMO} is applied to the moving object regions by randomly erasing their superpixels. 
	It enforces the network to pay attention to the foreground objects and learn the normal features more effectively, rather than simply predicting the future frame. 
	Moreover, \textit{MOLoss} urges the model to focus on learning normal objects captured within \textit{RandomSEMO} by amplifying the loss on the pixels near the moving objects.
	The experimental results show that our model outperforms state-of-the-arts on three benchmarks. 
	%Furthermore, extensive experiments well validate the effectiveness of \textit{RandomSEMO} and \textit{MOLoss}.
	
\end{abstract}

\vspace{-0.3cm}
\section{Introduction}
\label{sec:intro}
\vspace{-0.2cm}
Video anomaly detection (VAD) is a computer vision task that discriminates abnormal behaviors within the captured scenes.
It is gaining interest due to the demanding cost on monitoring surveillance videos.
Because the occurrence of abnormal events is rare compared to the normal, which is known as a data imbalance problem, the datasets used to develop anomaly detection models consist of a normal training set and a test set containing both normal and abnormal frames~\cite{avenue, st, ped2}. 
Following this circumstance, most of the deep-learning based VAD models~\cite{ffp, mnad, integrate, hybrid, macho, fastano, stan} are trained to recognize the normal patterns during training, and detect the frames with the outlying patterns during testing. 
Some methods~\cite{ffp, mnad, integrate, fastano} utilize frame predicting autoencoders (AEs) which generate the successive frame from the input frames based on the learned normal patterns. 
These approaches discriminate abnormal scenes by the quality of the predicted output, assuming that the unusual frame is likely to be generated poorly. 
%Similarly, some models~\cite{hybrid, stan, memae} adopt reconstruction tasks to the AEs.  

\begin{figure}[t]
	\centering
	\includegraphics[width=0.8\linewidth]{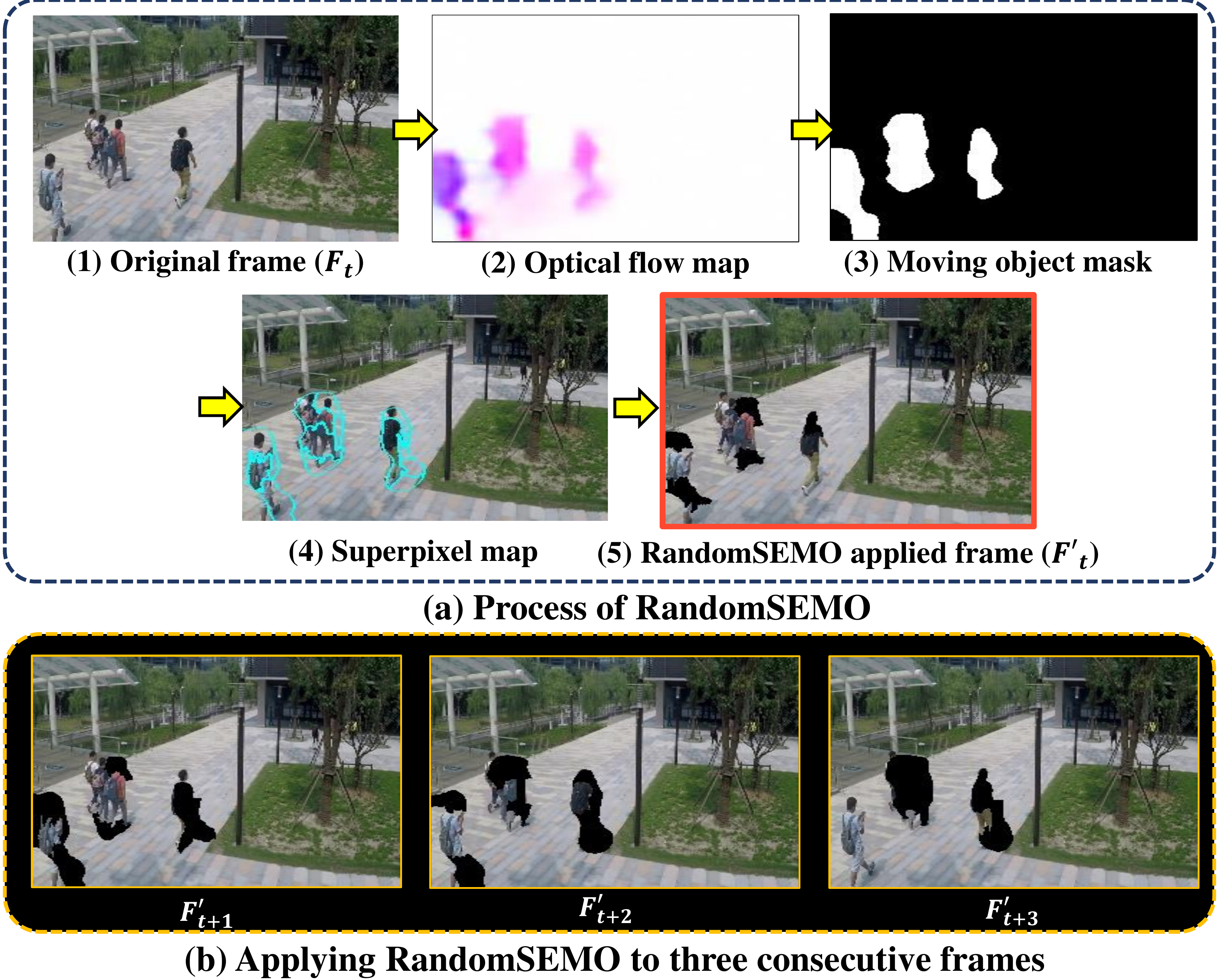}
	\caption{Process of the proposed RandomSEMO. (a) shows the process done for each frame. With the optical flow computed between $\mathbf{F_t}$ and $\mathbf{F_{t+1}}$ ((2)), we estimate the region of the moving objects ((3)). Then, on that region, we group pixels with similar characteristics to make superpixels ((4)). Finally, we erase some superpixels at a random probability ((5)). (b) illustrates application on three successive frames.}
	\label{fig:introduction}
	\vspace{-0.5cm}
\end{figure}

\begin{figure*}[!ht]
	\centering
	\includegraphics[width=0.75\textwidth]{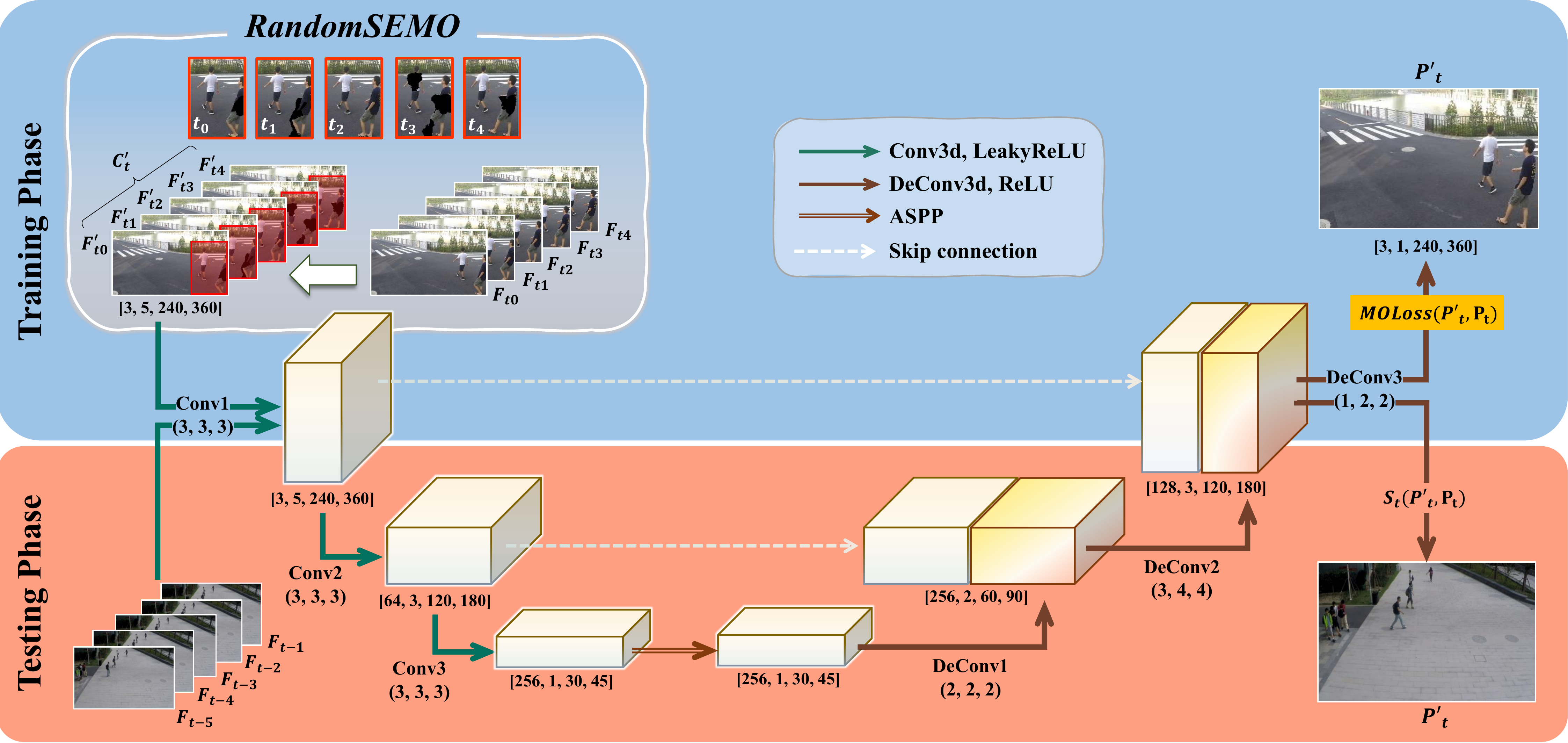}
	\caption{Overall framework of our method. During training, the AE predicts future frame $\mathbf{P'_t}$ from the RandomSEMO input frames $\mathbf{C'_t}$. The training exceeded with the help of the proposed MOLoss. When testing, the RandomSEMO is detached and normality score $S_t$ obtained from the quality of $\mathbf{P'_t}$ determines the normality of each frame. The values in brackets indicate [channel, temporal, height, width] of feature and (depth, height, width) of kernel in order.} 
	\label{fig:overall scheme}
	\vspace{-0.5cm}
\end{figure*}

However, these approaches face two major challenges. First, the AEs tend to perform excessively well, generating even the abnormal frames with high quality, which is adverse in detecting anomalies. 
Also, the large number of objects in the foreground and background distract the AE from focusing on the prominent moving objects. Abnormal behaviors generally take place in the moving foregrounds, rather than the background objects. 

To cope with these problems, inspired by the recently proposed data transformation algorithms~\cite{fastano, gold, joshi2019unsupervised}, we propose a novel video transformation technique named RandomSEMO.
It is applied in the data pre-processing stage before feeding the input to the frame predicting AE during training phase.
The RandomSEMO erases random superpixels found on the moving object regions. It works under the assumption that the model is likely to learn the most prominent objects' mormal patterns when it is given partially insufficient data for the frame predicting task. 

We also propose a Moving Object Loss (MOLoss) to maximize the effect of RandomSEMO. The MOLoss amplifies the loss of the pixels on the moving objects via generating a weight map. By utilizing the proposed loss function, our model is designed to focus on the foreground regions, effectively extracting crucial features with less distraction of the background objects. 

Our contributions are summarized as: (1) We propose a superpixel-based video transformation method called RandomSEMO. (2) We maximize the advantage of the RandomSEMO with the proposed loss function, MOLoss. (3) Our model surpasses or performs compatibly with other methods.

\vspace{-0.5cm}
\section{Related Works}
\label{sec:preliminary section}
\vspace{-0.3cm}
Various data transformation techniques, such as augmentations, have been proposed to boost the feature learning for computer vision tasks. In the image-level, Zhong {\textit{et al.}}~\cite{zhong2020random} introduced Random Erasing which makes occlusion at a random patch in the image. DeVries and Taylor~\cite{cutout} suggested CutOut, a method that deletes a box at the random location. These mentioned methods are all proven by experiments to enhance the performance of various tasks.

Few VAD methods have been proposed using the data transformation. Georgescu {\textit{et al.}}~\cite{georgescu2021anomaly} used sequence-reversed frames to embed motion feature learning. Park {\textit{et al.}}~\cite{fastano} devised SRT and TMT, which rotates random rectangular patches in the spatial dimension and shuffles the sequence of random patches, respectively. 
Our work is in close relation with this method. However, compared to the work of Park {\textit{et al.}}~\cite{fastano}, our proposed RandomSEMO is fully aware of the moving objects and applied only on the corresponding region. It transforms the superpixels of the region, which is more sophisticated and foreground-aware than applying transformations on random rectangular patches. Therefore, our method is more superior in the localizing ability, which is discussed in Sec.~\ref{sec:Qualitative Results} and Fig.~\ref{fig5:results}.

\vspace{-0.5cm}
\section{Method}
\label{sec:method}
\vspace{-0.3cm}
The overall framework is described in Fig.~\ref{fig:overall scheme}. We resize $N_f$ consecutive frames $\mathbf{F_{t-N_f}, F_{t-N_f+1}, F_{t-N_f+2}, \dots, F_{t-1}}$ to $3\times240\times360$ and stack them to make a 4D cuboid $\mathbf{C_t} \in \mathbb{R}^{3 \times N_f \times 240 \times 360}$. During training, RandomSEMO takes place to make  transform $\mathbf{C_t}$ to $\mathbf{C'_t}$. Then, $\mathbf{C'_t}$ is fed to the AE which generates a prediction of the future frame $\mathbf{P'_t}$. The AE is trained with the proposed MOLoss between $P'_t$ and the ground truth $P_t$. During inference, RandomSEMO is detached. $\mathbf{C_t}$ is fed to the AE, and the normality score $S_t$ is calculated from $\mathbf{P'_t}$ to distinguish the abnormal frames. 

\vspace{-0.5cm}
\subsection{RandomSEMO}
\vspace{-0.2cm}
We propose RandomSEMO, a transformation method that eliminates the details of the foregrounds by randomly erasing the superpixels of the moving objects. The purpose of RandomSEMO is to let AE be trained to extract prototypical normal features of the foreground objects than simply learning to predict frames well. 

To apply RandomSEMO, we first localize the moving objects. Given a sequence of frames $\mathbf{F_{t-N_f}}$, $\mathbf{F_{t-N_f+1}}$, $\mathbf{F_{t-N_f+2}}$, $\dots$, $\mathbf{F_{t-1}}$, $\mathbf{P_t}$, we compute the optical flow between each two successive frames (Fig.~\ref{fig:introduction} (2)). Then, we generate moving object masks $\mathbf{MoMask}$ (Fig.~\ref{fig:introduction} (3)) by thresholding the magnitude of the optical flow. Next, we apply SLIC~\cite{slic} on the masked regions of the former frames to generate superpixel maps (Fig.~\ref{fig:introduction} (4)) with maximum $N_{sp}$ components. Finally, we randomly erase the superpixels by replacing the values with zero (Fig.~\ref{fig:introduction} (5)). Each superpixel is erased with probability $T_{sp}$.

By utilizing RandomSEMO, the network is forced to learn the appearance and motion features of the moving objects to infer the future frame with the randomly erased information, as seen in Fig.~\ref{fig:introduction} (b). This solves the aforementioned excessive predicting power problem of the previous methods because our method makes the model to focus on the normal feature learning, rather than simply implying the upcoming frame with the cues of the input frames. Also, the AE effectively extracts normal features because RandomSEMO nudges the model to pay attention to the moving objects which are more potentially abnormal than the background.  

\begin{figure}[!t]
	\centering
	\includegraphics[width=1.0\columnwidth]{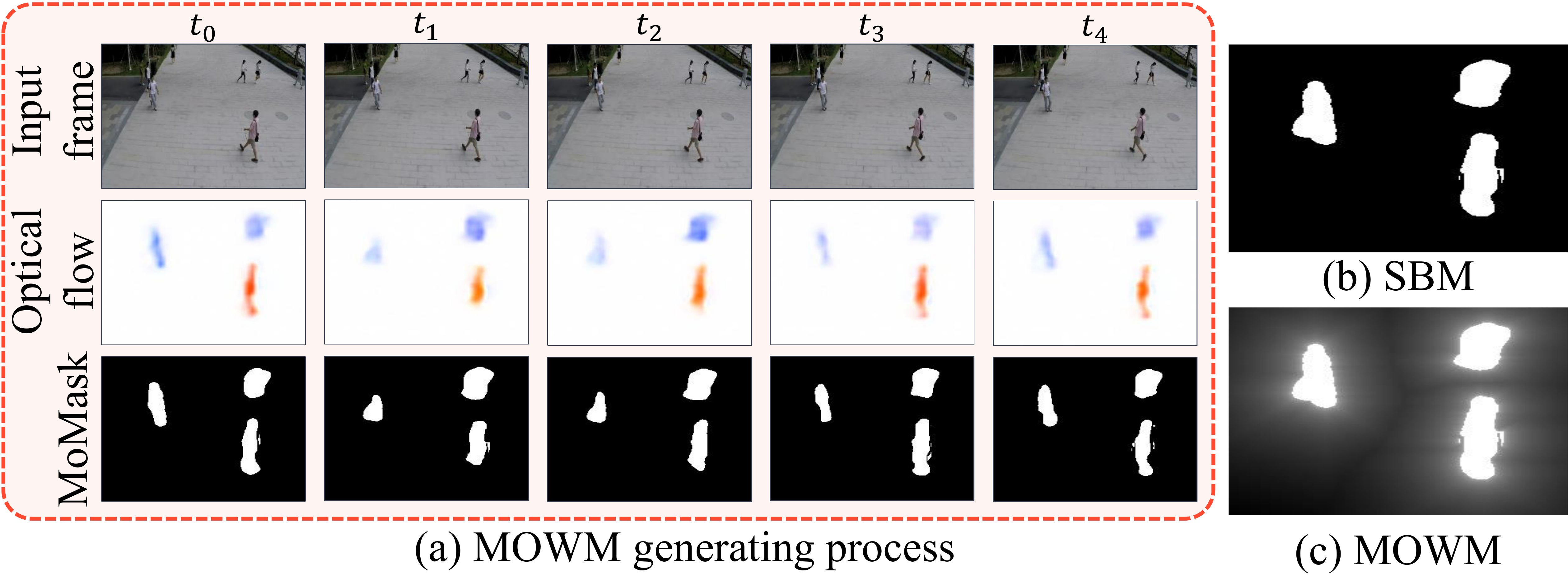}
	\caption{MOWM generating process.}
	\label{fig:mowm}
	\vspace{-0.5cm}
\end{figure}

\vspace{-0.4cm}
\subsection{MOLoss}
\vspace{-0.2cm}
To maximize the effect of RandomSEMO, we optimize our model with the newly proposed MOLoss. MOLoss is a weighted sum of a moving object weighted L1 loss $L_m$ and SSIM loss~\cite{ssim} $L_s$ obtained between $\mathbf{P'_t}$ and $\mathbf{P_t}$. 
$L_m$ amplifies loss on the pixels near the moving objects by using a Moving Object Weight Map ($\mathbf{MOWM}$). 
Fig.~\ref{fig:mowm} shows the example and generating process of MOWM. 
We sum up all the $\mathbf{MoMask}$ obtained during RandomSEMO to generate a summed binary mask ($\mathbf{SBM}$) (Fig.~\ref{fig:mowm} (b)), which represents the moving object region of the input frame cuboid $C'_t$. 
Therefore, $\mathbf{SBM}$ is expressed as:
\vspace{-0.2cm}
\small
\begin{equation}
	\mathbf{SBM _ { t }} = min\left(\sum _ { k=0 } ^ { t-1 } \mathbf{MoMask _ { k }}, 255\right)
	\vspace{-0.2cm}
\end{equation}
\normalsize
Then, we apply a distance function to $\mathbf{SBM}$ to generate $\mathbf{MOWM}$, which demonstrates the inverse-distance from each pixel to the foreground. Thereby, the values near the foreground pixel are large and vice versa. $\mathbf{MOWM}$ is defined as:

\vspace{-0.3cm}
\scriptsize
\begin{equation}
	\mathbf{MOWM^t} = 255-\begin{cases} min \sqrt { (p _ { x }-q _ { x } ) ^ { 2 } +(p _ { y }-q _ { y } ) ^ { 2 } } & , p \in \mathbf{SBM} _ { bg }  \\ \tab[2.0cm]0 & , p \in \mathbf{SBM} _ { fg } \end{cases} ,
\end{equation}
\normalsize
where $p$ and $q$ indicate each pixel in $\mathbf{SBM_t}$ and its nearest foreground pixel, respectively. 
Evidently, the pixels near the foreground have relatively big values for the corresponding location in $\mathbf{MOWM}$.  
Then, $\mathbf{MOWM}$ is multiplied during L1 loss calculation, amplifying the values near the moving objects. During this stage, we add a small value $\eta$ to the tensorized $\mathbf{MOWM}$ to avoid zero values. Because $L_m$ is a pixel-level loss function, we combine this with $L_s$ to guarantee the similarity of $\mathbf{P'_t}$ and $\mathbf{P_t}$ in the feature-level. The equations are as follows:

\vspace{-0.3cm}
\tiny
\begin{equation}
	\label{mol1}
	L_m(\mathbf{P'_t}, \mathbf{P_t}) = \frac{ 1 } { M \times N } \sum _ { x=0 } ^ { M-1 } \sum _ { y=0 } ^ { N-1 } \left[\left(\mathbf{MOWM} _ { (x,y) }^t+\eta\right) \times |\mathbf{P '} _ { (x,y) } - \mathbf{P} _ { (x,y) }|\right]
\end{equation}
\vspace{-0.3cm}
\begin{equation}
	\label{ssim}
	L_{s}(\mathbf{P'_t}, \mathbf{P_t}) = 1 - \frac{ (2 \mu _ { \mathbf{P'_t} } \mu _ { \mathbf{P_t} } + c _ { 1 } )(2 \sigma _ { \mathbf{P'_t}\mathbf{P_t} } + c _ { 2 } ) } { (2 \mu _ { \mathbf{P'_t} } ^ { 2 } \mu _ { \mathbf{P_t} } ^ { 2 } + c _ { 1 } )( \sigma _ { \mathbf{P'_t} } ^ { 2 } + \sigma _ { \mathbf{P_t} } ^ { 2 } + c _ { 2 } ) } 
\end{equation}
\normalsize
\noindent In Eq.~\ref{mol1}, M and N represent the number of pixels in the width and height axis, respectively. 
Moreover, $x$ and $y$ represent the location of each pixel in the width and height axis. In Eq.~\ref{ssim}, $\mu$ and $\sigma^2$ denote the average and variance of each frame, respectively. Also, $\sigma _ { \mathbf{P'_t}\mathbf{P_t} }$ represents the covariance. $c_1$ and $c_2$ are the variables to stabilize the division. Finally, MOLoss is described as:

\vspace{-0.4cm}
\small
\begin{equation}
	\label{MOLoss}
	MOLoss(\mathbf{P'_t}, \mathbf{P_t}) = w_mL_m(\mathbf{P'_t}, \mathbf{P_t}) + w_sL_{s}(\mathbf{P'_t}, \mathbf{P_t})
\end{equation}
\normalsize
\vspace{-0.5cm}

\noindent where $w_m$ and $w_s$ denote the weights that control the contribution of $L_m$ and $L_{s}$, respectively. 
%Please refer to the supplement for detailed descriptions and equations of the MOLoss and MOWM. 

\vspace{-0.4cm}
\subsection{AE Architecture}
\vspace{-0.2cm}
The AE used in our method is trained to recognize the likelihood of the normal frames and predict the upcoming frame $\mathbf{P'_t}$ from the transformed input frame cuboid $\mathbf{C'_t}$ using the learned normal features. To avoid excessive generalization, we design a lightweight, yet effective AE.  As demonstrated in Fig.~\ref{fig:overall scheme}, our AE is composed of a three-layer encoder and a three-layer decoder with skip connections in between to supplement the features lost during downsampling. Each encoder layer consists of 3D convolution, batch normalization, and LeakyReLU activation. After the last encoder layer, an Atrous Spatial Pyramid Pooling (ASPP)~\cite{aspp} layer takes place to enlarge the receptive field. The decoder layers are all made of 3D deconvolution, batch normalization, and ReLU activation. 

\vspace{-0.4cm}
\subsection{Normality Score}
\vspace{-0.2cm}
During inference, we adopt the peak signal to noise ratio (PSNR) to estimate the abnormality of each frame. It is defined as
%\vspace{-0.15cm}
%\begin{equation}
$PSNR(\mathbf{P'_t}, \mathbf{P_t}) = 10\log_{10} \frac{\max (\mathbf{P '_t})}{\|\mathbf{P'_t}-\mathbf{P_t}\|_2^2/N}$,
%\end{equation}
where N is the number of pixels in the frame. When $\mathbf{P_t}$ contains anomalies which our network has never seen during training, our network is incapable of predicting a clear $\mathbf{P'_t}$, resulting in low PSNR and vice versa. Following~\cite{st, memae, ffp,hybrid, mnad, stan, fastano, macho}, we normalize $PSNR(\mathbf{P'_t}, \mathbf{P_t}$) of each video clip to the range $[0, 1]$ to obtain the final normality score $S_t$.
\vspace{-0.15cm}
\small
\begin{equation}
	\label{score}
	S_t = \frac{PSNR(\mathbf{P'_t}, \mathbf{P_t}) - \min PSNR(\mathbf{P'_t}, \mathbf{P_t})} {\max PSNR(\mathbf{P'_t}, \mathbf{P_t}) - \min PSNR(\mathbf{P'_t}, \mathbf{P_t})}
\vspace{-0.15cm}
\end{equation}
\normalsize

\begin{table*}[t!]
	\centering
	\resizebox{0.9\textwidth}{!}{%
		\begin{tabular}{c|c|c|c|c|c|c|c|c||c|c|c}
			\hline
			\hline
			\diagbox[width=10em]{Dataset}{Model}& \multicolumn{1}{c|}{STAN~\cite{stan}} & \multicolumn{1}{c|}{\makecell{Abn\\GAN~\cite{abngan}}} & \multicolumn{1}{c|}{HyAE~\cite{hybrid}}  & \multicolumn{1}{c|}{\makecell{Mem\\AE~\cite{memae}}} & \multicolumn{1}{c|}{AD~\cite{ad}} & \multicolumn{1}{c|}{IntAE~\cite{integrate}} & \multicolumn{1}{c|}{\makecell{MNAD\\-Recon~\cite{mnad}}}  & \multicolumn{1}{c||}{\makecell{Fast\\Ano~\cite{fastano}}}  & \multicolumn{1}{c|}{Base}  & \multicolumn{1}{c|}{Base+R}  & \multicolumn{1}{c}{\makecell{\textbf{Base+R+M}\\ \textbf{(Ours)}}}\\ \hline \hline
%			\backslashbox{Dataset}{Model} & \multicolumn{1}{c|}{STAN~\cite{stan}} & \multicolumn{1}{c|}{\makecell{Abn\\GAN~\cite{abngan}}} & \multicolumn{1}{c|}{HyAE~\cite{hybrid}}  & \multicolumn{1}{c|}{\makecell{Mem\\AE~\cite{memae}}} & \multicolumn{1}{c|}{AD~\cite{ad}} & \multicolumn{1}{c|}{IntAE~\cite{integrate}} & \multicolumn{1}{c|}{\makecell{MNAD\\-Recon~\cite{mnad}}}  & \multicolumn{1}{c||}{\makecell{Fast\\Ano~\cite{fastano}}}  & \multicolumn{1}{c|}{Base}  & \multicolumn{1}{c|}{Base+R}  & \multicolumn{1}{c}{\makecell{\textbf{Base+R+M}\\ \textbf{(Ours)}}}\\ \hline \hline
			Avenue~\cite{avenue} & 81.7 & - & 82.8 & 83.3 & - & 83.7 & 82.8 &{\textbf{\textcolor{blue}{85.3}}} & 84.1 & 83.9 & {\textbf{\textcolor{red}{85.4}}} \\
			ST~\cite{st} & - & - & - & 71.2 & - & 71.5 & 69.8 &{\textbf{\textcolor{blue}{72.2}}} & 71.6 & {\textbf{\textcolor{blue}{72.2}}} & {\textbf{\textcolor{red}{72.4}}} \\
			Ped2~\cite{ped2} & 92.2 & 93.5 & 84.3 & 94.1 & 95.5 & {\textbf{\textcolor{blue}{96.2}}} &90.2 & {\textbf{\textcolor{red}{96.3}}} & 93.7 & 94.2 & 95.8 \\
			\hline
			FPS & 50 & - & - & 38 & 2 & 30 & 67 & {\textbf{\textcolor{red}{195}}} & {\textbf{\textcolor{blue}{127}}} & {\textbf{\textcolor{blue}{127}}} & {\textbf{\textcolor{blue}{127}}} \\
			\hline
			\hline
		\end{tabular}%
	}
	\vspace{-0.1cm}
	\caption{Frame-level AUC (\%) comparison. All figures and FPS are copied from the corresponding papers. 'Base' indicates our baseline framework without the RandomSEMO and MOLoss. It is instead trained with a single $L_1$ loss. R and M represent RandomSEMO and MOLoss, respectively. The top two results for each category are marked {\textbf{\textcolor{red}{red}}} and {\textbf{\textcolor{blue}{blue}}}.}
	\vspace{-0.4cm}
	\label{tab:model comparison}
\end{table*}

\vspace{-0.3cm}
\section{Experiments}
\label{sec:experiment}
\vspace{-0.3cm}
\noindent \textbf{Dataset.} We validate our network on three popular benchmarks: CUHK Avenue~\cite{avenue}, ShanghaiTech Campus (ST)~\cite{st}, and UCSD Ped2~\cite{ped2}. These datasets are all acquired by fixed cameras from the real-world. Among the three, ST~\cite{st} is the largest and most complex, containing 330 training videos and 107 testing videos captured from 13 different scenes. 

\noindent \textbf{Evaluation metric.} For the evaluation, we adopt the area under curve (AUC) obtained from the frame-level scores and the ground truth labels. This metric is used in most studies~\cite{ffp, hybrid, integrate, fastano, memae, gold} on VAD. Since the first five frames of each clip cannot be predicted, they are ignored in the evaluation, following other prediction-based methods~\cite{ffp, integrate, mnad, fastano}.

\noindent \textbf{Settings.} We implement our experiments using PyTorch and a NVIDIA RTX A6000. Also, we use a pre-trained FlowNet2~\cite{flownet} to generate optical flow maps. The training is based on batch size $30$ and Adam optimizer with an initial learning rate of $2e^{-4}$, $\beta_1 = 0.5$, $\beta_2 = 0.9$. The network is trained for 40 epochs on Ped2~\cite{ped2} and Avenue~\cite{avenue} and $10$ epochs for ST~\cite{st}. $\eta$, $w_m$, $w_s$, $N_{f}$, $N_{sp}$, and $T_{sp}$ are empirically set to $1, 0.25, 0.75, 5, 10$, and $0.3$ respectively.

%\begin{table}[t!]
%	\centering
%	\resizebox{0.5\columnwidth}{!}{%
%		\begin{tabular}{c|c|c}
%			\hline
%			\hline
%			& Avenue~\cite{avenue}   & Ped2~\cite{ped2}    \\ \hline\hline
%			$T_{sp}=0.1$ & $84.1$ & $94.9$ \\
%			$T_{sp}=0.2$ & $85.0$ & $\textbf{95.9}$ \\
%			$T_{sp}=0.3$ & $\textbf{85.4}$ & $95.8$ \\
%			$T_{sp}=0.4$ & $84.5$ & $95.0$ \\
%			$T_{sp}=0.5$ & $84.4$ & $94.1$ \\
%			$T_{sp}=0.6$ & $84.4$ & $94.7$ \\
%			\hline\hline     
%		\end{tabular}%
%	}
%	\caption{Impact of the RandomSEMO probability $T_{sp}$}
%	\label{tab:ablation1}
%	\vspace{-0.3cm}
%\end{table}

\vspace{-0.4cm}
\subsection{Quantitative Results}
\label{ssec:quantitative result}
\vspace{-0.2cm}
The quantitative results are shown in Table~\ref{tab:model comparison}. We compare our network with eight VAD networks that do not require any pre-trained network—such as object detectors or optical flow networks—during inference for fair comparison. It is observed that the accuracy of our model is superior to most of the compared methods in the three datasets. 
%Especially, our network surpasses all others on ST~\cite{st}. This reveals that learning the likelihood of the moving objects without distraction of the non-salient background is useful for datasets with various scenes. 
Furthermore, our proposed network is capable of detecting anomalies at 127 frames per seconds (FPS), faster than most methods. Because accuracy and speed are both mandatory in the real world application, our network is more practical than other methods that show low accuracy or slow speed.

\begin{figure}[t]
	\begin{minipage}{\textwidth}
		\begin{minipage}[b]{0.24\textwidth}
			\centering
			\resizebox{1\textwidth}{!}{%
				\begin{tabular}{c|c|c}
					\hline
					\hline
					& Avenue~\cite{avenue}   & Ped2~\cite{ped2}    \\ \hline\hline
					$T_{sp}=0.1$ & $84.1$ & $94.9$ \\
					$T_{sp}=0.2$ & $85.0$ & $\textbf{95.9}$ \\
					$T_{sp}=0.3$ & $\textbf{85.4}$ & $95.8$ \\
					$T_{sp}=0.4$ & $84.5$ & $95.0$ \\
					$T_{sp}=0.5$ & $84.4$ & $94.1$ \\
					$T_{sp}=0.6$ & $84.4$ & $94.7$ \\
					\hline\hline     
				\end{tabular}%
			}
			\captionof{table}{Impact of the RandomSEMO probability $T_{sp}$}
		\label{tab:ablation1}
		\end{minipage}
		\begin{minipage}[b]{0.24\textwidth}
			\centering
			\includegraphics[width=0.8\textwidth]{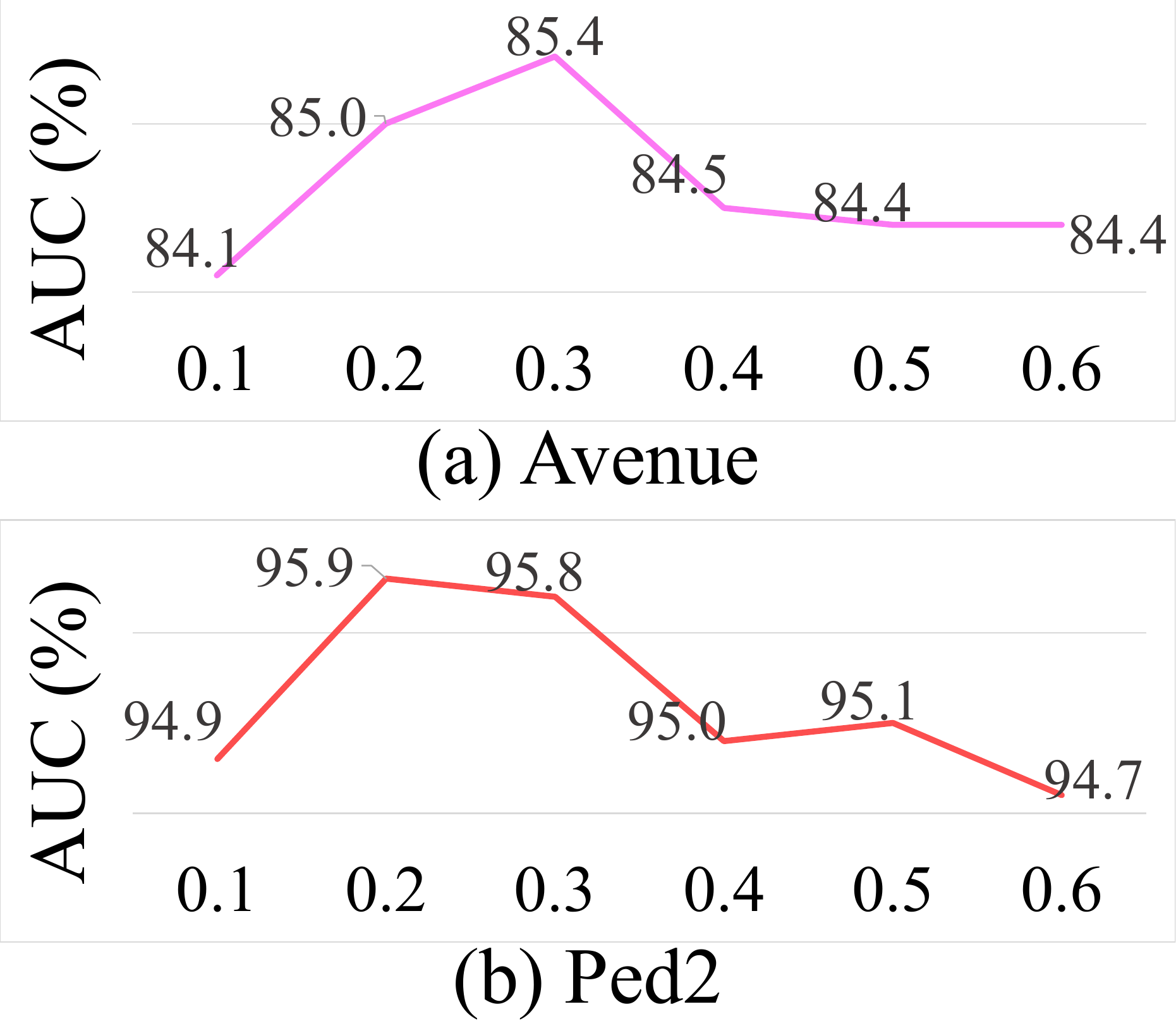}
			\captionof{figure}{Graph of Table \ref{tab:ablation1}}
		\label{fig:ablation1}
		\end{minipage}
	\end{minipage}
%	\captionlistentry[table]{Impact of the RandomSEMO probability $T_{sp}$}
%	\captionsetup{labelformat=andtable}
	\vspace{-0.2cm}
%	\caption{Impact of RandomSEMO probability}
\end{figure}

\begin{table}[t!]
	\centering
	\resizebox{0.7\columnwidth}{!}{%
		\begin{tabular}{c|c|c|c}
			\hline
			\hline
			& Avenue~\cite{avenue}   & ST~\cite{st}  & Ped2~\cite{ped2}    \\ \hline\hline
			$L_1$ & $83.9$ &$72.2$ & $94.2$\\
			SSIM~\cite{ssim} & $84.3$ & $72.0$ & $94.2$ \\
			$w_mL_1 + w_s$SSIM~\cite{ssim} & $84.8$ & $72.3$ & $94.2$ \\
			\textbf{MOLoss} & $\textbf{85.4}$ & $\textbf{72.4}$ & $\textbf{95.8}$ \\ \hline \hline    
		\end{tabular}%
	}
	\caption{Impact of MOLoss. Comparison with $L_1$, SSIM~\cite{ssim}, and a weighted sum of $L_1$ and SSIM~\cite{ssim}.}
	\label{tab:ablation2}
	\vspace{-0.5cm}
\end{table}

\vspace{-0.5cm}
\subsection{Qualitative Results}
\label{sec:Qualitative Results}
\begin{figure}[!t]
	\centering
	\includegraphics[width=0.8\columnwidth]{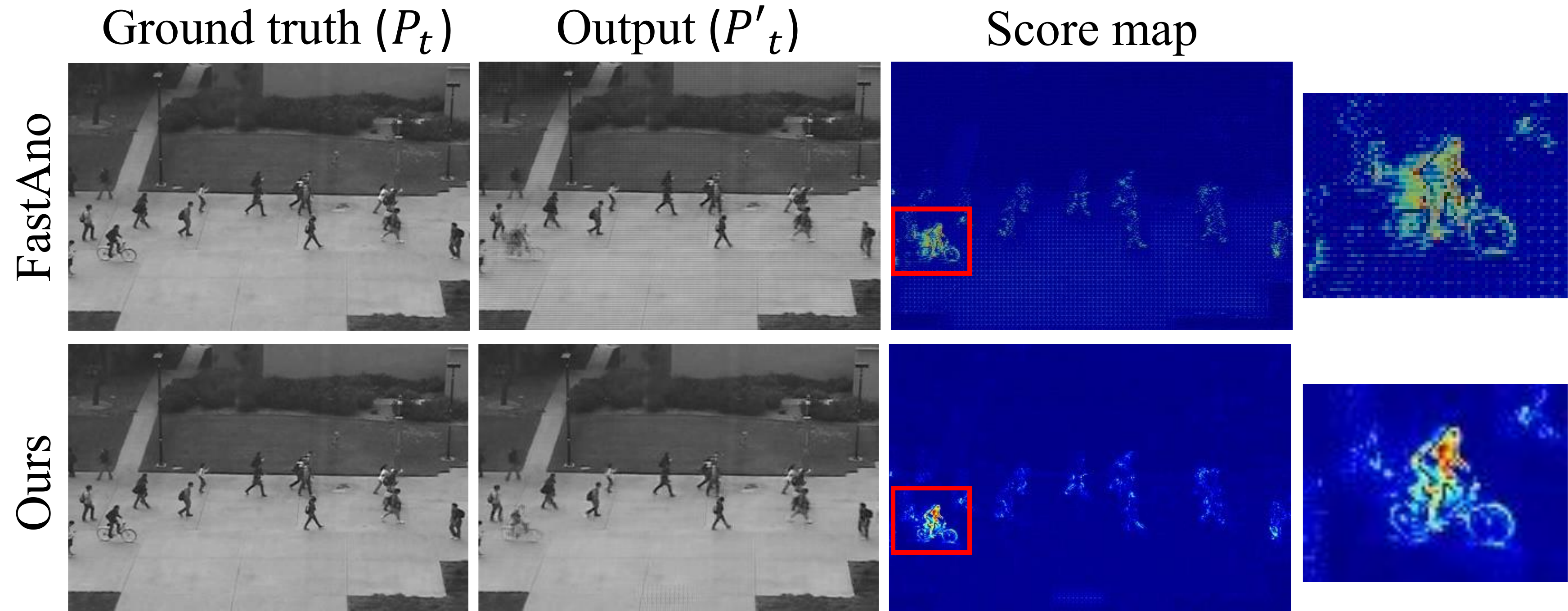}
	\caption{Qualitative results on Ped2~\cite{ped2} compared with those of FastAno~\cite{fastano}. The biker on the pedestrian street is the anomaly in this scene.}
	\label{fig5:results}
	\vspace{-0.5cm}
\end{figure}

\vspace{-0.2cm}
Fig.~\ref{fig5:results} demonstrates the output and difference map of our network compared to those of FastAno~\cite{fastano}. The difference map shows that our network more precisely localizes the abnormal object. Only the biker is saturated in our proposed model whereas the background pixels near the biker is unnecessarily highlighted in FastAno~\cite{fastano}, mirroring the effect of our moving object aware RandomSEMO and MOLoss. Furthermore, the difference between foreground and background is more pronounced in our results. 

\vspace{-0.5cm}
\subsection{Impact of RandomSEMO probability $T_{sp}$}
\label{sec:ablation study1}
\vspace{-0.2cm}
We found the optimal value of $T_{sp}$ by changing $T_{sp}$ from $0.1$ to $0.6$. Table~\ref{tab:ablation1} and Fig.~\ref{fig:ablation1} show the result of the experiments conducted on Avenue~\cite{avenue} and Ped2~\cite{ped2}. For Avenue~\cite{avenue}, the accuracy is highest when $T_{sp}$ is $0.3$. For Ped2~\cite{ped2}, the model trained with $T_{sp}=0.2$ shows the best performance, and it is also as much accurate when trained with $T_{sp}=0.3$. Furthermore, it is observed that both low $T_{sp}$ leads to relatively low performance because only a few superpixels are obscured, practically having no significant difference from the original image. Similarly, the accuracy drops when $T_{sp}$ is high because too much information is lost.

\vspace{-0.4cm}
\subsection{Impact of MOLoss}
\label{sec:ablation study2}
\vspace{-0.2cm}
We conducted ablation studies to validate the contribution of MOLoss. 
The results are compared in Table~\ref{tab:ablation2}.
We experimented by changing the loss function to $L_1$ loss, SSIM loss, and a weighted sum of $L_1$ and SSIM where the weights are equivalent to those of the MOLoss. From the results, it is observed that using MOLoss strongly boosts the performance, especially when compared to using a single $L_1$ loss. 

\vspace{-0.5cm}
\section{Conclusion}
\label{sec:conclution}
\vspace{-0.3cm}
We propose a novel superpixel-based video transformation method called \textit{RandomSEMO} for anomaly detection. 
Also, we propose a loss function named \textit{MOLoss} to boost the effect of \textit{RandomSEMO} by amplifying the loss values on the pixels near the moving objects. 
%Thereby, our model effectively captures the normal features during training by focusing on the moving foreground. 
The experimental results have shown that our network performs competitively or surpasses other previous methods. 
Furthermore, we validated the impact of \textit{RandomSEMO} and \textit{MOLoss} with ablation studies and output comparison. 
As future works, we will verify the localizing ability of our model for extensive usage.

\vfill
\pagebreak
\newpage
\begingroup
\setstretch{0.9}
\bibliographystyle{IEEEbib}
\bibliography{strings,refs}

\begin{thebibliography}{10}

\bibitem{avenue}
Cewu Lu, Jianping Shi, and Jiaya Jia,
\newblock ``Abnormal event detection at 150 fps in matlab,''
\newblock in {\em ICCV}, 2013, pp. 2720--2727.

\bibitem{st}
Weixin Luo, Wen Liu, and Shenghua Gao,
\newblock ``A revisit of sparse coding based anomaly detection in stacked rnn
  framework,''
\newblock in {\em ICCV}, 2017, pp. 341--349.

\bibitem{ped2}
Vijay Mahadevan, Weixin Li, Viral Bhalodia, and Nuno Vasconcelos,
\newblock ``Anomaly detection in crowded scenes,''
\newblock in {\em CVPR}. IEEE, 2010, pp. 1975--1981.

\bibitem{ffp}
Wen Liu, Weixin Luo, Dongze Lian, and Shenghua Gao,
\newblock ``Future frame prediction for anomaly detection--a new baseline,''
\newblock in {\em CVPR}, 2018, pp. 6536--6545.

\bibitem{mnad}
Hyunjong Park, Jongyoun Noh, and Bumsub Ham,
\newblock ``Learning memory-guided normality for anomaly detection,''
\newblock in {\em CVPR}, 2020, pp. 14372--14381.

\bibitem{integrate}
Yao Tang, Lin Zhao, Shanshan Zhang, Chen Gong, Guangyu Li, and Jian Yang,
\newblock ``Integrating prediction and reconstruction for anomaly detection,''
\newblock {\em PR Letters}, vol. 129, pp. 123--130, 2020.

\bibitem{hybrid}
Trong~Nguyen Nguyen and Jean Meunier,
\newblock ``Hybrid deep network for anomaly detection,''
\newblock {\em BMVC}, 2019.

\bibitem{macho}
MyeongAh Cho, Taeoh Kim, Ig-Jae Kim, and Sangyoun Lee,
\newblock ``Unsupervised video anomaly detection via normalizing flows with
  implicit latent features,''
\newblock {\em arXiv preprint arXiv:2010.07524}, 2020.

\bibitem{fastano}
Chaewon Park, MyeongAh Cho, Minhyeok Lee, and Sangyoun Lee,
\newblock ``Fastano: Fast anomaly detection via spatio-temporal patch
  transformation,''
\newblock in {\em WACV}, 2022, pp. 2249--2259.

\bibitem{stan}
Sangmin Lee, Hak~Gu Kim, and Yong~Man Ro,
\newblock ``Stan: Spatio-temporal adversarial networks for abnormal event
  detection,''
\newblock in {\em ICASSP}. IEEE, 2018, pp. 1323--1327.

\bibitem{memae}
Dong Gong, Lingqiao Liu, Vuong Le, Budhaditya Saha, Moussa~Reda Mansour, Svetha
  Venkatesh, and Anton van~den Hengel,
\newblock ``Memorizing normality to detect anomaly: Memory-augmented deep
  autoencoder for unsupervised anomaly detection,''
\newblock in {\em ICCV}, 2019, pp. 1705--1714.

\bibitem{gold}
Muhammad~Zaigham Zaheer, Jin-ha Lee, Marcella Astrid, and Seung-Ik Lee,
\newblock ``Old is gold: Redefining the adversarially learned one-class
  classifier training paradigm,''
\newblock in {\em CVPR}, 2020, pp. 14183--14193.

\bibitem{joshi2019unsupervised}
Abhishek Joshi and Vinay~P Namboodiri,
\newblock ``Unsupervised synthesis of anomalies in videos: transforming the
  normal,''
\newblock in {\em IJCNN}. IEEE, 2019, pp. 1--8.

\bibitem{zhong2020random}
Zhun Zhong, Liang Zheng, Guoliang Kang, Shaozi Li, and Yi~Yang,
\newblock ``Random erasing data augmentation,''
\newblock in {\em AAAI}, 2020, vol.~34, pp. 13001--13008.

\bibitem{cutout}
Terrance DeVries and Graham~W Taylor,
\newblock ``Improved regularization of convolutional neural networks with
  cutout,''
\newblock {\em arXiv preprint arXiv:1708.04552}, 2017.

\bibitem{georgescu2021anomaly}
Mariana-Iuliana Georgescu, Antonio Barbalau, Radu~Tudor Ionescu, Fahad~Shahbaz
  Khan, Marius Popescu, and Mubarak Shah,
\newblock ``Anomaly detection in video via self-supervised and multi-task
  learning,''
\newblock in {\em CVPR}, 2021, pp. 12742--12752.

\bibitem{slic}
Radhakrishna Achanta, Appu Shaji, Kevin Smith, Aurelien Lucchi, Pascal Fua, and
  Sabine S{\"u}sstrunk,
\newblock ``Slic superpixels,''
\newblock Tech. {R}ep., 2010.

\bibitem{ssim}
Hang Zhao, Orazio Gallo, Iuri Frosio, and Jan Kautz,
\newblock ``Loss functions for image restoration with neural networks,''
\newblock {\em IEEE Transactions on computational imaging}, vol. 3, no. 1, pp.
  47--57, 2016.

\bibitem{3dconv}
Du~Tran, Lubomir Bourdev, Rob Fergus, Lorenzo Torresani, and Manohar Paluri,
\newblock ``Learning spatiotemporal features with 3d convolutional networks,''
\newblock in {\em ICCV}, 2015, pp. 4489--4497.

\bibitem{leakyrelu}
Andrew~L Maas, Awni~Y Hannun, Andrew~Y Ng, et~al.,
\newblock ``Rectifier nonlinearities improve neural network acoustic models,''
\newblock in {\em ICML}. Citeseer, 2013, vol.~30, p.~3.

\bibitem{aspp}
Liang-Chieh Chen, George Papandreou, Iasonas Kokkinos, Kevin Murphy, and Alan~L
  Yuille,
\newblock ``Deeplab: Semantic image segmentation with deep convolutional nets,
  atrous convolution, and fully connected crfs,''
\newblock {\em TPAMI}, vol. 40, no. 4, pp. 834--848, 2017.

\bibitem{relu}
Vinod Nair and Geoffrey~E Hinton,
\newblock ``Rectified linear units improve restricted boltzmann machines,''
\newblock in {\em ICML}, 2010.

\bibitem{flownet}
Eddy Ilg, Nikolaus Mayer, Tonmoy Saikia, Margret Keuper, Alexey Dosovitskiy,
  and Thomas Brox,
\newblock ``Flownet 2.0: Evolution of optical flow estimation with deep
  networks,''
\newblock in {\em CVPR}, 2017, pp. 2462--2470.

\bibitem{adam}
Diederik~P Kingma and Jimmy Ba,
\newblock ``Adam: A method for stochastic optimization,''
\newblock {\em arXiv preprint arXiv:1412.6980}, 2014.

\bibitem{abngan}
Mahdyar Ravanbakhsh, Moin Nabi, Enver Sangineto, Lucio Marcenaro, Carlo
  Regazzoni, and Nicu Sebe,
\newblock ``Abnormal event detection in videos using generative adversarial
  nets,''
\newblock in {\em ICIP}. IEEE, 2017, pp. 1577--1581.

\bibitem{ad}
Mahdyar Ravanbakhsh, Enver Sangineto, Moin Nabi, and Nicu Sebe,
\newblock ``Training adversarial discriminators for cross-channel abnormal
  event detection in crowds,''
\newblock in {\em WACV}. IEEE, 2019, pp. 1896--1904.

\bibitem{dualgan}
Fei Dong, Yu~Zhang, and Xiushan Nie,
\newblock ``Dual discriminator generative adversarial network for video anomaly
  detection,''
\newblock {\em IEEE Access}, vol. 8, pp. 88170--88176, 2020.

\end{thebibliography}
\endgroup

\end{document}